\pdfoutput=1

\documentclass[11pt]{article}

\usepackage[]{acl}

\usepackage{times}
\usepackage{latexsym}
\usepackage{amsmath}
\usepackage{amsfonts}
\usepackage[ruled]{algorithm2e}
\usepackage[T1]{fontenc}
\usepackage{subcaption}
\usepackage{graphicx}

\usepackage[utf8]{inputenc}

\usepackage{microtype}
\usepackage{colortbl}
\usepackage{booktabs}
\usepackage{tabu}

\title{Misconfidence-based Demonstration Selection for LLM In-Context Learning}

\author{Shangqing Xu \and Chao Zhang \\
  Georgia Institute of Technology \\
  \texttt{\{sxu452, chaozhang\}@gatech.edu}\\}

\begin{document}
\maketitle

\begin{abstract}

  In-context learning with large language models (LLMs) excels at adapting to various tasks rapidly.
  However, its success hinges on carefully selecting demonstrations, which remains an obstacle in practice.
  Current approaches to this problem either rely on hard-to-acquire external supervision or require frequent interactions with LLMs, resulting in high costs.
  We propose a new method called In-Context Reflection (ICR) to overcome these challenges.
  ICR strategically selects demonstrations to reduce the discrepancy between the LLM's outputs and the actual input-output mappings.
  Specifically, ICR starts with a random set of initial demonstrations, then iteratively refines it.
  In each step, it analyzes a pool of candidate examples and identifies the ones most likely to challenge the LLM's current understanding, measured by a new metric called misconfidence.
  These most confusing examples are then selected to replace the less informative demonstrations in the current set.
  Our comprehensive evaluation across five diverse datasets encompassing 13 subtasks shows the efficacy of ICR.
  Compared to existing methods, ICR achieves an average performance boost of 4\%, while demonstrating remarkable cross-task generalization capabilities.

\end{abstract}

\section{Introduction}

In-context learning (ICL, \citet{brown2020language}) enables pre-trained large language models (LLMs) to adapt to diverse tasks by appending question-answer pairs (demonstrations) as prompt contexts. Despite its effectiveness, ICL can be highly sensitive to the quality of the demonstrations \citep{zhao2021calibrate, min-etal-2022-rethinking}, emphasizing the need for strategies to strategically select ICL demonstrations.

\begin{figure}
  \centering
  \includegraphics[width = \linewidth]{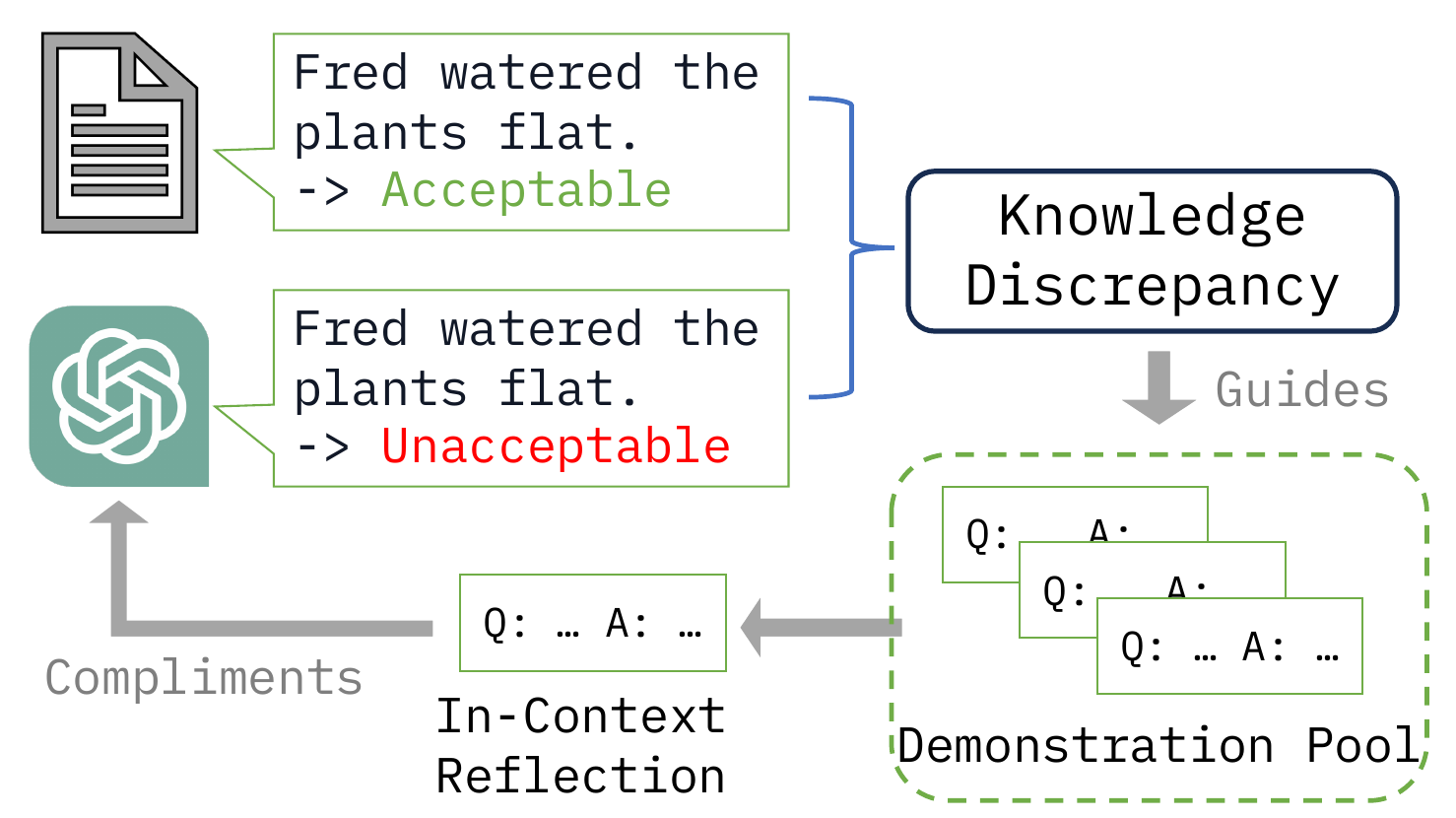}
  \caption{A overview of our method. We aim to leverage the exact discrepancy between LLM's knowledge and task input-output mappings. Then we select demonstrations that best bridge such discrepancies.}
  \label{fig:intro_small}
\end{figure}

Existing demonstration selection strategies roughly fall into two categories. One approach first obtains external supervision through preferences of pre-trained encoders or retrievers, then adopts a scorer to assign scores for each demonstration candidate based on the supervision. Such scorers could be a semantic distance model \citep{liu-etal-2022-makes, gao2023ambiguityaware}, a reward function \citep{rubin-etal-2022-learning, zhang2022active}, or a reversed topic predictor \citep{wang2023large}. The other approach estimates the importance of each candidate by influence analysis, which contrasts LLM predictions before and after adding the candidate to the prompt. The influence can be computed via task-agnostic measures, such as mutual information gain \citep{sorensen-etal-2022-information} or validation performance gain \citep{li2023finding, nguyen2023context}.

While these approaches have shown promising performance in selecting demonstrations for ICL, they suffer from the following limitations.
On the one hand, adopting a scorer depends on reliable external supervision to score accurately.
As in-context learning's mechanism is different from such external supervision, the demonstrations prioritized by the supervision may not be the best choice for forming in-context prompts.
Additionally, these methods often require fine-tuning another LLM, which can be computationally expensive.
On the other hand, influence analysis by contrast needs one to perform a large number of binary tests with the LLM, which is costly and unscalable for handling large numbers of candidates.

Our idea is to directly leverage the discrepancy between the output distribution of LLMs and task-specific input-output mappings.
This discrepancy arises when the LLM assigns high probabilities to incorrect labels.
By constructing ICL prompts that bridge these discrepancies, we aim to calibrate the LLM's output distribution toward the desired task labels.
This strategy is effective because it directly addresses the influence of demonstrations on the LLM through ICL.
It is also efficient, as it eliminates the need for repeated binary tests for contrasting.

We present In-Context Reflection (ICR), a new method for selecting effective ICL demonstrations from a certain pool based on LLM \textit{misconfidence}. First, we approximate the aforementioned discrepancies by obtaining LLM's predictions for each candidate in the pool based on an initial prompt. Candidates that are more confidently misjudged
by the LLM (that is, candidates with higher misconfidence) indicate gaps between LLM's distribution and task mappings, and are therefore prioritized.  Consequently, we re-rank all candidates based on their misconfidence and replace the initial prompt with top-ranked ones to construct a refined prompt.

To validate the effectiveness of our method, we conducted experiments across five diverse task sets, encompassing 13 distinct tasks ranging from sentiment analysis to complex language comprehension challenges. Our analysis demonstrates that the prompts generated using our method achieve an average performance improvement of 4\% across all tasks. This shows that ICR consistently enhances the LLM's performance across these tasks. Furthermore, to measure the robustness of ICR, we generate prompts for one dataset and subsequently evaluate them in the same task family. We found that different-task ICR is comparable to same-task uniform sampling, highlighting its potential for broad applications.

Our main contribution includes:

\begin{itemize}

  \item 
We propose leveraging the difference between the output distribution of LLMs and the input-output mappings of a given task to address the drawbacks of existing demonstration selection strategies.

  \item 
We introduce misconfidence as a metric to quantify this discrepancy and present In-Context Reflection (ICR), a method that effectively selects demonstrations that provide "lacking knowledge" to help LLMs adapt to specific tasks.

  \item
Through experiments on 13 tasks from 5 task sets, we demonstrate that prompts constructed using ICR are both effective and robust.

\end{itemize}

\section{Related Work}

In-context learning (ICL) \citep{brown2020language} empowers LLMs to rapidly adapt to a wide range of tasks. While ICL proves effective across English-based tasks \citep{min2021metaicl}
and multilingual tasks \citep{lin-etal-2022-shot}, it exhibits significant sensitivity to various factors, including prompt design \citep{lester-etal-2021-power}, demonstration distribution \citep{min-etal-2022-rethinking}, instruction design \citep{mishra2022reframing}, and demonstration ordering \citep{zhao2021calibrate, lu-etal-2022-fantastically}. Given these intricate dependencies, it's crucial to develop advanced demonstration selection.

Following the categorization proposed by \citet{dong2023survey}, we classify demonstration selection strategies into two categories: 1) adapting a task-specific scorer with external supervision to guide demonstration selection, and 2) contrast-based task-agnostic measures derived from the LLM's predictions.

\noindent \textbf{Learned  Scorers} Adapted scorers typically provide pairwise scores between each test case and the pool of candidate demonstrations. \citet{liu-etal-2022-makes} proposed to use Sentence-BERT \citep{izacard2021unsupervised} to generate semantical embeddings, and introduce k-Nearest Neighbors to pick demonstrations. \citet{gao2023ambiguityaware} further enhanced this approach by retrieving candidates whose ground label lies in top-2 zero-shot predictions. Further, \citet{rubin-etal-2022-learning} trained a GPT-Neo as a contrastive scorer as well as a demonstration referrer, and \citet{li2023unified} advanced this framework through unified training across various datasets. \citet{ye2023compositional} introduced Determinantal Point Processes (DPPs) to model the interaction between sequences of demonstrations, which enables retrieving a set of demonstrations.

On the other hand, some approaches are trying to obtain individual scores and build prompts that work for all test cases. \citet{zhang2022active} introduced Q-learning to train a retriever that could actively adapt to previously unseen tasks. \citet{wang2023large} fine-tuned a smaller LLM as a task-specific token encoder and ranked demonstrations according to their ability to rebuild tokens.

Some studies have also investigated the discrepancy between the output distribution of LLMs and the input-output mappings of tasks. For instance,
\citet{gao2023ambiguityaware} first calculate the zero-shot prediction of test cases then retrieve semantically closer candidates whose label lies in the top predictions. \citet{mavromatis2023examples} assume each wrongly-judged demonstration could mostly assist in judging cases from its semantic neighborhood, therefore formalizing demonstration selection as max coverage problem.
While these approaches share a similar methodology with ours, they rely on semantic distances whereas our method quantifies the misconfidence in LLM outputs.

\noindent \textbf{Contrasting Task-Agnostic Measures}
A straightforward approach involves randomly selecting demonstrations from the entire candidate pool, as suggested by
\citet{min-etal-2022-rethinking}. \citet{sorensen-etal-2022-information}
assessed a prompt sequence by calculating the mutual information between predicted outcomes and true labels. \citet{nguyen2023context} proposed constructing a validation set and evaluating each train instance by contrasting the validation performance of prompts with and without the instance. \citet{li2023finding} introduced InfoScore, a computationally efficient pipeline that iteratively filters train samples.
However, these methods still require conducting many tests with the LLM, which becomes prohibitively expensive and unscalable when dealing with a large number of candidates.

\section{Problem Formulation}

We investigate few-shot in-context learning (ICL) with pre-trained LLMs for specific tasks.
A target task comprises
a train set $\mathcal{D}_{train} = \{(x_i, y_i)\}_{i=1}^{N_{train}}$ 
and a test set $\mathcal{D}_{test} = \{(x_i, y_i)\}_{i=1}^{N_{test}}$. The data from both the train and test sets are independently and identically distributed (i.i.d), and the labels in the test set are not available except when doing evaluations. In this paper, we limit the task to a single-label classification task, assuming that all labels fall in certain categories $y_i \in Y_{\mathcal{D}},\  \forall (x_i, y_i) \in \mathcal{D} = \{\mathcal{D}_{train} \cup \mathcal{D}_{test}\}$.

We predict for the target task via few-shot ICL.
Given a pre-trained LLM $\theta$, ICL adapts $\theta$ towards a specific target task.
Given a input $x$ from $\mathcal{D}_{test}$, we concatenate original input $x$ with the demonstrations, changing the output probability into $p_\theta(y|x_1, y_1, \dots, x_n, y_n, x)$.
We denote the probability of $y$ given $x$ and demonstration $\mathcal{P}$ as $p_\theta(y|x, \mathcal{P})$.

The demonstrations must be carefully selected from a candidate pool
$\mathcal{C}\subseteq \mathcal{D}_{train}$, forming a subset
$\mathcal{P} = \{(x_j, y_j)\}_{j=1}^m \subset \mathcal{C}, \ m \ll N_{train}$.
 We assume that the model parameters
$\theta$
remain fixed throughout the process, and only the selection of
$\mathcal{P}$
is modified. The success of this adaptation is measured by the predictive accuracy, namely whether the output generated by the LLM, $y_\theta(x_i) = \mathrm{argmax}_{y \in Y_{\mathcal{D}}}p_\theta(y|x_I, \mathcal{P})$, matches the actual label $y_i$.
 The central challenge of demonstration selection is to identify the optimal subset
$\mathcal{P} \subset \mathcal{C}$ that yields the most significant improvement in prediction accuracy, measured as  $\mathrm{Acc}(y_i, y_{(\theta, \mathcal{P})}(x_i)),\ \forall (x_i, y_i) \in \mathcal{D}_{test}$.

\section{Method}

\begin{figure*}
  \centering
  \includegraphics[width = \linewidth]{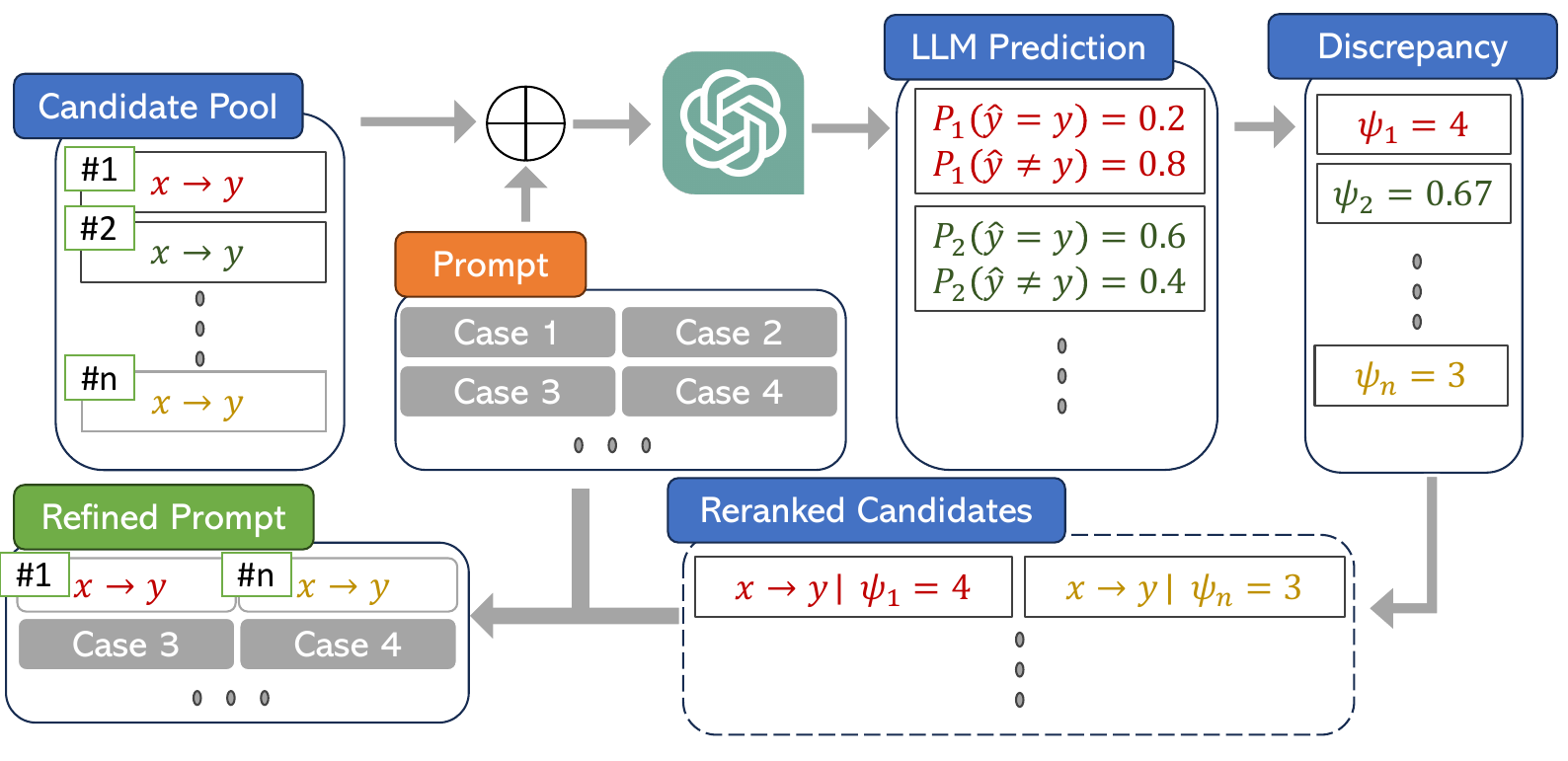}
  \caption{An overview of In-Context Reflection (ICR). We first use an initialized prompt to get LLM prediction for each candidate, then calculate the misconfidence score $\psi$ to measure the discrepancy between LLM and task. After that, we rerank all candidates according to $\psi$, and replace part of the prompt with the top-ranked candidates, obtaining a refined prompt.}
  \label{fig:method}
\end{figure*}

\subsection{In-Context Learning by Bridging Discrepancy}

Let us first define discrepancy in our ICR method. While trying to adapt a LLM $\theta$ for  $\mathcal{D}$, there can be cases where $\theta$'s output $y_\theta$ doesn't match the actual label
$y$. We say there is \textit{discrepancy} between LLM's output probability $p_\theta$ and task's input-output mappings $\{x_i \rightarrow y_i,\ (x_i, y_i) \in \mathcal{D}\}$. By minimizing such discrepancy, LLM would always output correct labels for each input, therefore adapting $\theta$ to $\mathcal{D}$.

The objective of ICL is equivalent to minimizing discrepancy. As shown by recent studies  \citep{zhao2021calibrate, min-etal-2022-rethinking, wei2023larger}, ICL's demonstrations $\mathcal{P}$ contribute by guiding $\theta$ to mimic mappings $\{x_i \rightarrow y_i, \ (x_i, y_i) \in \mathcal{P}\}$, and to generate output $y_{(\theta, \mathcal{P})}$ for $x$ correspondingly. As most ICL methods select $\mathcal{P}$ that are representative of $\mathcal{D}$ \citep{min-etal-2022-rethinking}, such ICL methods are therefore equal to bridge the discrepancy between $p_\theta$ and $\{x_i \rightarrow y_i, \ (x_i, y_i) \in \mathcal{D}\}$.

Now we propose how to select ICL demonstrations that can best bridge such discrepancies. Consider a prompt $\mathcal{P}$. If the discrepancy between $\{x_i \rightarrow y_i, \ (x_i, y_i) \in \mathcal{P}\}$ and $p_\theta$ is small, $\mathcal{P}$ should have limited influences on $p_\theta$ as its mappings are already obtained by $\theta$. In this case, mimicking $\{x_i \rightarrow y_i, \ (x_i, y_i) \in \mathcal{P}\}$ would barely change $p_\theta(y|x, \mathcal{P})$ from $p_\theta(y|x)$. This inference also holds reversely. Therefore, to bridge the discrepancies between $p_\theta$ and $\{x_i \rightarrow y_i, \ (x_i, y_i) \in \mathcal{D}\}$, we can select $\mathcal{P}$ whose mappings have the largest discrepancy between $p_\theta$.

We propose that the discrepancy between the set of input-output pairs \(\{(x_i, y_i)\}\) and the model's predictions \(p_\theta(x)\) can be approximated by the misconfidence associated with each case \((x_i, y_i)\) with respect to the model \(\theta\). Intuitively, if a case \((x_i, y_i)\) is easily misclassified by the model (i.e., \(p_\theta(x_i) \neq y_i\)), we consider the misconfidence of that case to be high. High misconfidence indicates that the model struggles to correctly predict the output label \(y_i\) for the input \(x_i\). Consequently, the overall discrepancy between the observed data \(\{(x_i, y_i)\}\) and the model's predictions \(p_\theta(x)\) is expected to be high.

We thus quantify the misconfidence of a model by measuring the margin between the highest probability assigned to any incorrect label, $\max_{y \in Y, y \neq y_i} p_\theta(y|x_i)$, and the output probability of the correct label, $p_\theta(y_i|x_i)$. This margin reflects how confidently the model misjudges the true label from plausible alternatives. We denote this score as $\psi((x_i, y_i), \theta)$ and compute it as:
\begin{align} \label{eq:psi}
  \psi((x_i, y_i), \theta) = \frac{\max_{y \neq y_i, y \in Y} p_\theta(y|x_i)}{p_\theta(y_i|x_i)}
\end{align}

Further, given an initial prompt $\mathcal{P}_0$, we can compute the probability conditioned on these demonstrations, which yields the misconfidence score:

\begin{align}
  \psi((x_i, y_i), (\theta, \mathcal{P}_0)) = \frac{\max_{y \neq y_i, y \in Y} p_\theta(y|x_i, \mathcal{P}_0)}{p_\theta(y_i|x_i, P_0)}
\end{align}

Such prompt-based misconfidence score helps us select candidates that can enhance $\mathcal{P}_0$. 

\subsection{In-Context Reflection (ICR)}\label{sec:ICR}

To effectively adapt the model parameter $\theta$ to a specific task, we select demonstrations based on their $\psi$ scores (Equation \ref{eq:psi}). We introduce the In-Context Reflection (ICR) pipeline, which uses this strategy to efficiently construct an optimal demonstration set $\hat{\mathcal{P}}$.

The ICR pipeline begins with an initial demonstration set $\mathcal{P}_0 = \{(x_1, y_1), \dots (x_m, y_m)\}$, which is randomly sampled from the candidate pool $\mathcal{C}$. In each iteration, ICR updates the misconfidence score for all candidates based on the current demonstration set. Then, it reranks the candidates according to their misconfidence and replaces $n$ of the previous demonstration set with these top-ranked candidates. 
The entire process is provided in Algorithm \ref{alg:icr}, where $+$($-$) denotes set-wise merging (excluding). Instead of solely re-ranking, we use iterative replacement to build the prompt, which has been shown crucial for obtaining a semantic distribution from $\mathcal{D}$ \citep{min-etal-2022-rethinking}.  Note that, ICR only requires one interaction with the LLM per train case, making it computationally efficient.

\begin{algorithm}[hbt!]
  \caption{In-Context Reflection}\label{alg:icr}
  \KwData{$\mathrm{LLM}\ \theta, \mathrm{Candidate\ Pool}\ \mathcal{C},$ \\ Demonstration size $m$, Replacing Number $n$ \\ Iteration Number $k$\\}
  \KwResult{$\mathrm{Optimal \ Prompt}\ \hat{\mathcal{P}}  $}
  $\mathcal{P}_0 = \phi$\;
  \For{$i = 1$ \KwTo m}{ 
    Sample $(x, y) \sim U(\mathcal{D})$\\
    $\mathcal{P}_0 = \mathcal{P}_0 + \{(x, y)\}$\\
  }
  $\mathcal{C} = \mathcal{C} - \mathcal{P}_0$\;
  \For{$i = 0$ \KwTo $k-1$}{
    \For{$(x,y) \in \mathcal{C}$}{
      Calculate $\psi((x, y), (p_\theta, \mathcal{P}_i))$\\
    }
    Rerank $(x,y) \in \mathcal{C}$ according to $\psi$\;
    $\mathcal{P}_{i+1} = \mathcal{C}[1:n] + \mathcal{P}_{i}[n+1:m]$\;
    Add the replaced to pool $\mathcal{C} = (\mathcal{C} - \mathcal{P}_{i+1}) + \mathcal{P}_{i}$\;
  }
  $\hat{\mathcal{P}} = \mathcal{P}_{k}$\;
\end{algorithm}

\section{Experiment}
\subsection{Settings}

\subsubsection{Datasets}

We evaluate ICR on 5 task sets containing 13 binary or multi-class classification tasks, detailed as follows. 
For GLUE, Ethos, and TweetEval, we only select part of the tasks, as other tasks contain too many test cases or are too easy for our backbone LLM to solve. Details are shown in Appendix \ref{sec:unselected}.

\noindent \textbf{GLUE} \citep{wang-etal-2018-GLUE} A multiple-task generalization benchmark covering topics from hypothesis to fact-checking. We adopt four subtasks: \textit{MRPC}, \textit{WNLI}, \textit{COLA}, \textit{RTE}.

\noindent \textbf{Ethos} \citep{Mollas_2022} A collection of hate speech detection tasks from online texts. We adopt four subtasks: \textit{Religion, Race, Gender, Directed\_vs\_generalized}

\noindent \textbf{TweetEval} \citep{barbieri-etal-2020-tweeteval} A multiple-task benchmark built from Twitter, all framed as multi-class classification. We adopt three subtasks: \textit{hate, emotion, irony}

\noindent \textbf{HateSpeech18} \citep{de-gibert-etal-2018-hate} A binary-labeled hate speech dataset extracted from a white supremacist forum.

\noindent \textbf{Poem Sentiment} \citep{sheng-uthus-2020-investigating} A multi-class sentiment dataset of poem verses from Project Gutenberg.

\subsubsection{Baselines}

We compare with the following baselines:

\noindent \textbf{Uniform Sampling} \citep{min-etal-2022-rethinking} We uniformly sample demonstrations stratifying the origin label distribution from the full candidate pool.

\noindent \textbf{Best-of-10} \citep{zhang2022active} We randomly sample 10 sets of demonstrations and select the best one by evaluating on a 100 validation subset.

\noindent \textbf{Topic} \citep{wang2023large} We fine-tune a GPT-2-Large-774M model to encode task-specific latent concept tokens, then select demonstrations whose in-context prompts best predicted the concept. We only adopt this method on GLUE, Ethos, and PoemSentiment, as there are no concept data available for other tasks.

\noindent \textbf{KATE} \citep{liu-etal-2022-makes} We introduce a pre-trained SBERT \citep{reimers-2019-sentence-bert} to calculate semantic embeddings for both candidate pool and test set. Then, for each input case from the test set, we retrieve the k-nearest neighbors from the candidate pool as the demonstrations.

\noindent \textbf{AMBIG} \citep{gao2023ambiguityaware}. For each test case, we perform zero-shot prediction and identify the labels in the top two predictions as the `Ambiguity label'. We then filter candidates with ground labels matching the Ambiguity label and choose semantically similar demonstrations from this subset.

\subsection{Evaluation and Implementation Details}

We use GPT-3.5-Turbo-Instruct as the backbone and use the same prompt format for all the methods.
Appendix \ref{sec:app_dataset} shows the task details as well as prompt formats.
By default, we select 16 demonstrations for all methods.

In each task, we use the full train set as the candidate pool for 1) Uniform Sampling, 2) KATE 3) AMBIG 4) Best-of-10, but we restrict Topic's candidate pool to a uniformly selected subset with a size of 500 to cut computational cost. To provide a fair comparison, we also restrict ICR's pool samely.
Evaluation is applied on test sets, except GLUE, where we evaluate methods on the validation set as there is no publicly available test label. We calculate both the macro-average F1 score and the accuracy score.

On all the tasks, we only adopt exactly one iteration of ICR and set the replacement count $n$ as 8, meaning that we will replace 8 demonstrations out of the original prompt by misconfidence re-ranking. We will provide results for multiple iterations in section \ref{sec:ab_iteration}.

\subsection{Same-Task Evaluation}

We show the same-task evaluation results in Table \ref{exp:main}. All the reported scores are the average of three independent runs with different random seeds, except for KATE and AMBIG, as they do not involve any randomness. For Ethos, TweetEval, and GLUE, as they involve multiple subtasks, we report the average score of the subtasks.

The results show that ICR outperforms all baselines (both scorer-based and contrasting-based) on all tasks with significant $4\%$ improvements. Our method solely relies on the candidate pool and corresponding LLM judgment, without employing any external knowledge base. Such an improvement shows the efficacy of leveraging discrepancies while building in-context prompts. 

In contrast, AMBIG performs poorly on all tasks compared to vanilla KATE, not to mention ICR. While it was claimed that this method can effectively capture clues from the LLM's distribution, the experimental results indicate that it relies more heavily on the SBERT semantic encoder (we will discuss this further in section \ref{sec:ab_geometrical}). Conversely, ICR with an originally designed $\psi$ score achieved much better results, showing that it can measure the discrepancy between LLM distribution and the task labels more accurately.

\begin{table*}
  \centering
  \begin{tabular}{lcccccc}
    \toprule
    & \multicolumn{6}{c}{Macro-F1} \\ \cline{2-7}
    & GLUE & Ethos & TweetEval & HateS18 & Poem & Average \\ \hline
    Uniform & $75.5$ & $65.5$ & $63.7$ & $63.7$ & $68.7$ & \cellcolor[HTML]{FEE4E3}$67.4$ \\
    Best-of-10 \citep{zhang2022active} & $75.8$ & $69.1$ & $68.8$ & $70.6$ & $72.2$ & \cellcolor[HTML]{FEE4E3}$71.3$ \\
    Topic \citep{wang2023large} & $76.2$ & $62.4$ & - & - & $75.2$ & \cellcolor[HTML]{FEE4E3}- \\
    KATE \citep{liu-etal-2022-makes} & $72.3$ & $71.2$ & $66.3$ & $66.8$ & $73.2$ & \cellcolor[HTML]{FEE4E3}$69.9$ \\
    AMBIG \citep{gao2023ambiguityaware} & $76.6$ & $71.3$ & $68.3$ & $72.4$ & $67.7$ & \cellcolor[HTML]{FEE4E3}$71.3$ \\ \hline
    \rowcolor[HTML]{D3FBFA}
    ICR (ours) & $\mathbf{78.7}$ & $\mathbf{76.5}$ & $\mathbf{71.0}$ & $\mathbf{74.4}$ & $\mathbf{76.5}$ & $\mathbf{75.4}$ \\ \midrule
    \multicolumn{1}{l}{} & \multicolumn{6}{c}{Accuracy} \\ \cline{2-7}
    \multicolumn{1}{l}{} & GLUE & Ethos & TweetEval & HateS18 & Poem & Average \\ \hline
    Uniform & $76.6$ & $70.7$ & $64.5$ & $74.2$ & $70.2$ & \cellcolor[HTML]{FEE4E3}$71.2$ \\
    Best-of-10 \citep{zhang2022active} & $77.3$ & $75.3$ & $69.8$ & $82.6$ & $74.0$ & \cellcolor[HTML]{FEE4E3}$75.8$ \\
    Topic \citep{wang2023large} & $77.6$ & $70.1$ & - & - & $76.9$ & \cellcolor[HTML]{FEE4E3}- \\
    KATE \citep{liu-etal-2022-makes} & $73.4$ & $76.7$ & $68.1$ & $78.2$ & $76.0$ & \cellcolor[HTML]{FEE4E3}$74.5$ \\
    AMBIG \citep{gao2023ambiguityaware} & $78.0$ & $76.5$ & $69.3$ & $83.8$ & $76.0$ & \cellcolor[HTML]{FEE4E3}$76.0$ \\ \hline
    \rowcolor[HTML]{D3FBFA}
    ICR (ours) & $\mathbf{80.6}$ & $\mathbf{82.2}$ & $\mathbf{71.6}$ & $\mathbf{87.0}$ & $\mathbf{78.9}$ & $\mathbf{80.0}$ \\ \bottomrule
  \end{tabular}
  \caption{
    \label{exp:main} Results on each task set. ICR outperforms all baselines with an average $4\%$ gain. It is an exciting result, as ICR uses no fine-tuning data and requires linearly scaled interactions with LLM.
  }
\end{table*}

\subsection{Different-Task Evaluation}

\begin{figure}[h]
  \centering
  \includegraphics[width = 0.6 \linewidth]{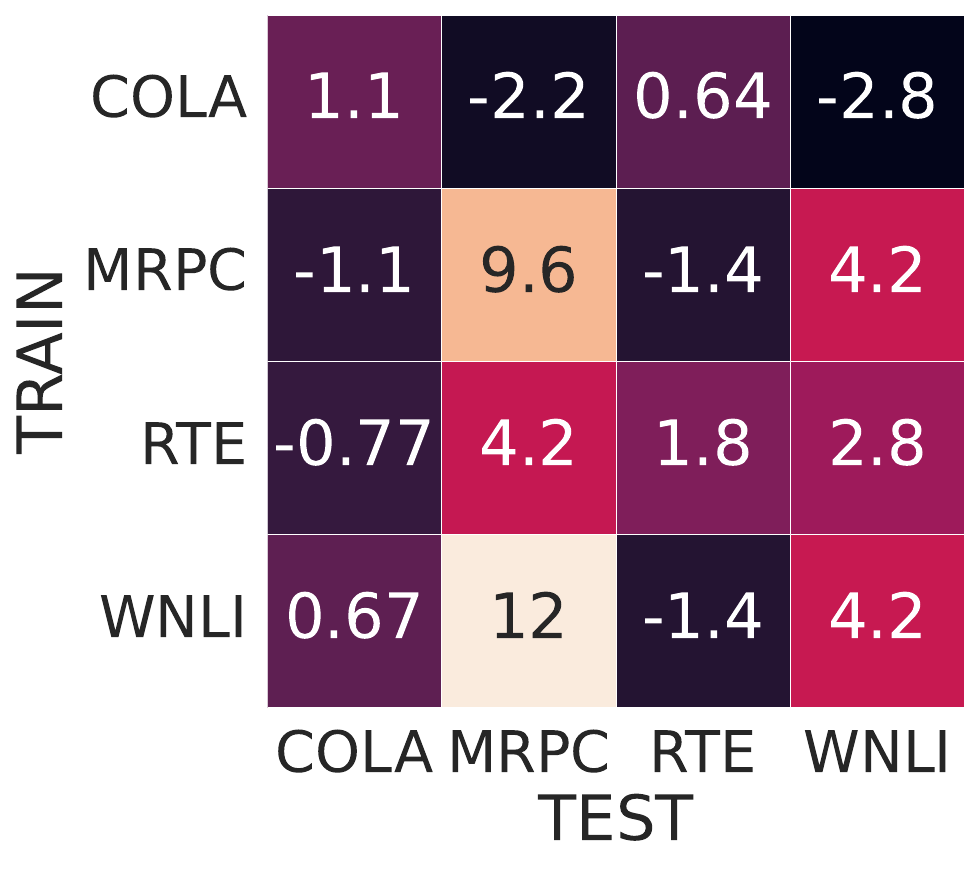}
  \caption{Different-task evaluation accuracy of ICR's prompt on GLUE's tasks. Each number shows the performance gain compared with same-task uniform sampling. On all tasks, ICR received comparable (sometimes even superior) results.
  }
  \label{fig:intra_GLUE}
\end{figure}

To further test the robustness of ICR, we build ICR prompts on each task of GLUE and evaluate them on different tasks from the same task family. We compare them to uniform prompts created for the same task. Figure \ref{fig:intra_GLUE} shows the result. Even when selecting demonstrations from different task sets, ICR obtains comparable (or sometimes even superior) results compared with same-task uniform sampling prompts. Also, we notice that the performance gains
between MRPC and WNLI are much higher than the rest, implying some latent correlations between these tasks.

\subsection{Ablation}

\subsubsection{Multiple ICR Iterations} \label{sec:ab_iteration}

\begin{figure}[h]
  \begin{subfigure}{\linewidth}
    \centering
    \includegraphics[width = 0.47\linewidth]{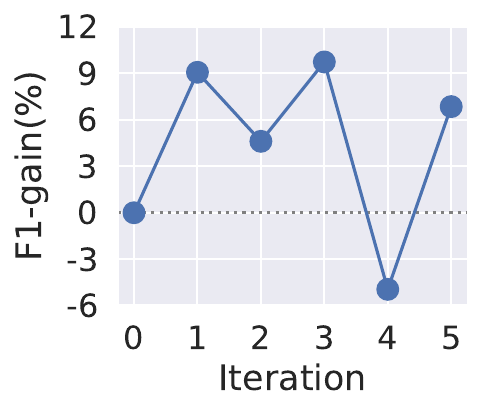}
    \hfill
    \includegraphics[width=0.47\linewidth]{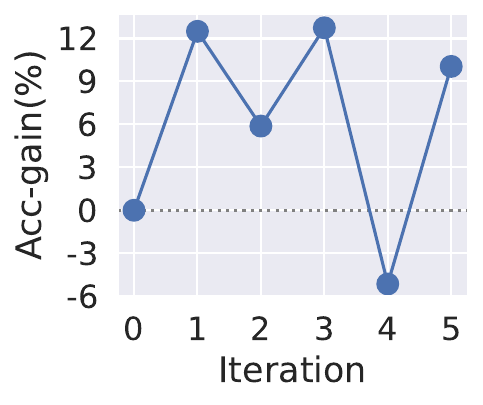}
    \caption{GLUE-MRPC}
  \end{subfigure}%

  \begin{subfigure}{\linewidth}
    \centering
    \includegraphics[height=0.36\linewidth]{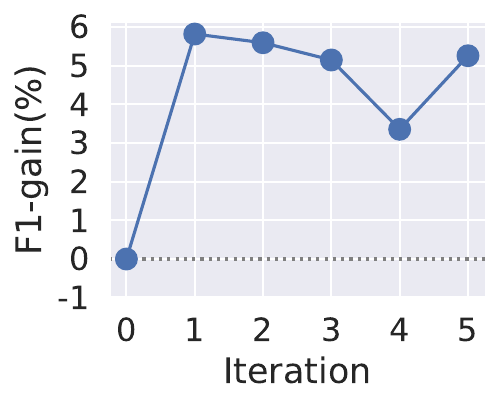}
    \hfill
    \includegraphics[height=0.355\linewidth]{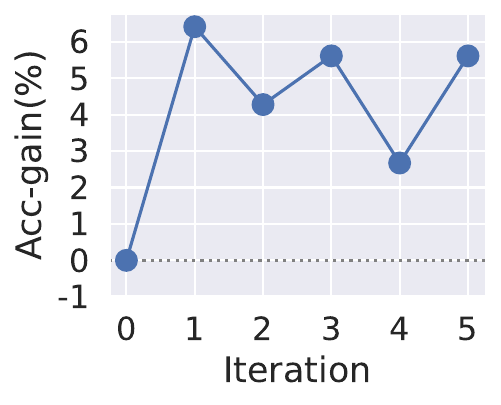}
    \caption{TweetEval-Emotion}
  \end{subfigure}%

  \caption{Performance gain of ICR on random prompt for 5 iterations. ICR produces a positive effect most times but is unstable. }
  \label{fig:ab_iteration}
\end{figure}

We investigate whether multiple ICR iterations yield better results. We initialize a random prompt and apply ICR iteratively for 5 iterations. Results on GLUE-MRPC and TweetEval-Emotion are presented in Figure \ref{fig:ab_iteration}. While ICR always contributes positively, each iteration does not consistently improve the performance. 
One possible reason is that in each iteration, the ICR update is too rough and large, causing the result to fluctuate around the global optimum. Therefore, we choose to set the number of iterations to one.

\subsubsection{Relationship between Misconfidence and Performance} \label{sec:ab_confusability}

\begin{figure}[h]
  \subfloat[GLUE-MRPC]{\includegraphics[height = 0.35\linewidth]{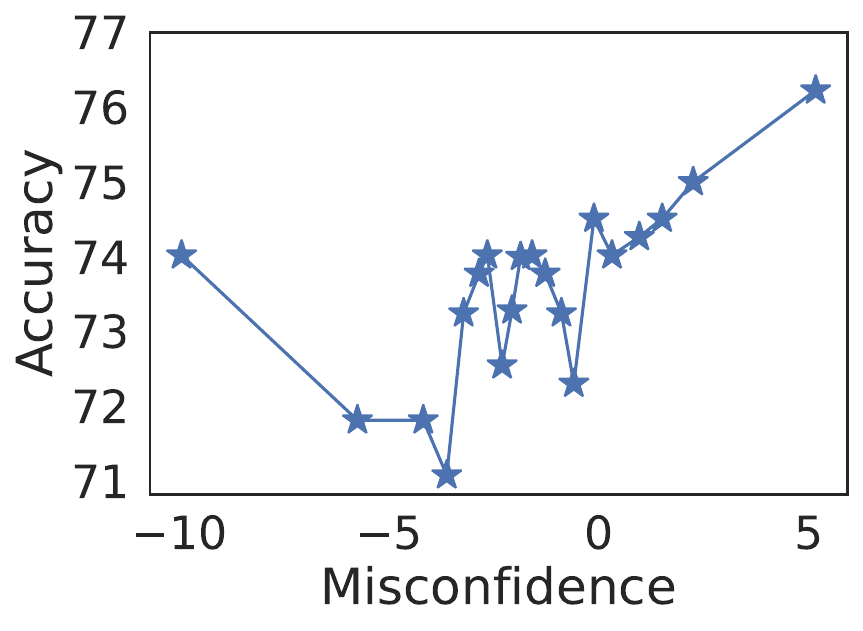}}
  \subfloat[Poem Sentiment]{\includegraphics[height = 0.35\linewidth]{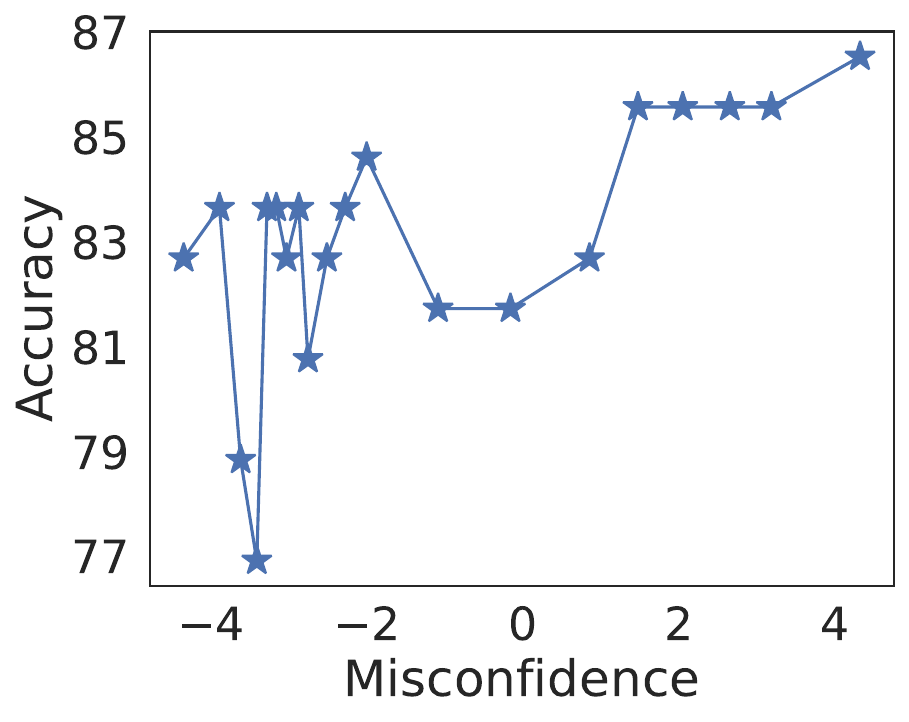}}
  \caption{Visualization of the result from demonstrations with different misconfidence levels on a) GLUE-MRPC b) Poem Sentiment. Demonstrations with larger misconfidence lead to better in-context performance.}
  \label{fig:ab_confusability}
\end{figure}

One key rationale for ICR is that demonstrations with larger misconfidence lead to better contribution. To confirm this idea, we build prompts with demonstrations of different misconfidence averages on a) Poem Sentiment and b) GLUE-MRPC, and evaluate them on the same task. The result is shown in Figure \ref{fig:ab_confusability}. We see performance of prompts generally is consistent with the demonstrations' misconfidence average. Also, it is interesting that the demonstrations with extremely low misconfidence (that is, they are correctly judged confidently) show better contributions than borderline ones. It shows that LLM can also be enhanced by further distinguishing confident knowledge.

\subsubsection{Initialization and Prompt Building}

As described in Section \ref{sec:ICR}, our algorithm first calculates misconfidence through a few-shot prompt, then does replacements to build a more powerful one. It is different from most similar studies, where they calculate measurements (or scores) by zero-shot prompts and build refined prompts entirely from such measurements. We therefore introduce two corresponding ablations.

\noindent \textbf{Initialization} We update the misconfidence using a zero-shot prompt instead of the original few-shot prompt. Then we build ICR prompts and compare their performance to the original ones. The result is shown in Table \ref{tab:ab_zeroshot}. We see the performance drops severely in most task sets. Given the result in section \ref{sec:ab_confusability}, we conclude that few-shot prompts can provide more reasonable misconfidence estimations, which makes ICR perform better.

\begin{table}[h]
  \begin{tabular}{l|lllll}
    \hline
    & GLUE & Ethos & TEval & HS18 & Poem \\ \hline
    F1 & $-5.1$ & $-10.0$ & $-5.6$ & $-6.0$ & $-6.1$ \\
    Acc & $-4.4$ & $-9.9$ & $-4.9$ & $-7.6$ & $-6.7$ \\ \hline
  \end{tabular}
  \caption{Performance change of ICR using zero-shot misconfidence instead of original few-shot ones. Using zero-shot misconfidence leads to significant drawbacks.}
  \label{tab:ab_zeroshot}
\end{table}

\noindent \textbf{Prompt Building} We select all demonstrations according to misconfidence instead of replacing part of initialized prompts. The result is shown in Table \ref{tab:ab_fullconfu}. Referring solely to misconfidence leads to a significant performance drop, except on GLUE. Note that, GLUE's labels are well-balanced, but other tasks' are not. Recalling conclusions from \citet{min-etal-2022-rethinking}, we see such performance drop is caused by a lack of label distribution information in the prompt. Therefore, building ICR prompt through partial replacement maintains label distribution information, and therefore is better than building solely from consufability re-ranking.

\begin{table}[h]
  \begin{tabular}{c|ccccc}
    \hline
    & GLUE & Ethos & TEval & HS18 & Poem \\ \hline
    F1 & $-1.7$ & $-7.9$ & $-6.8$ & $-10.4$ & $-7.7$ \\
    Acc & $-0.6$ & $-10.4$ & $-5.1$ & $-12.0$ & $-8.7$ \\ \hline
  \end{tabular}
  \caption{Performance change of ICR by selecting demonstration entirely from misconfidence instead of replacement. This leads to performance drops on all task sets except GLUE, as GLUE is highly balanced and therefore less affected by demonstration distribution.}
  \label{tab:ab_fullconfu}
\end{table}

\subsubsection{Influence of Semantic Distances} \label{sec:ab_geometrical}

In addition to ICR, several studies \citep{mavromatis2023examples, gao2023ambiguityaware} have suggested selecting demonstrations jointly based on LLM's output distribution and semantic distances. For instance, \citet{mavromatis2023examples} propose a method where they assume that borderline candidates have a strong influence on their semantic neighbors. However, when selecting ICL demonstrations, semantic distances and distribution measurements (like misconfidence) are independent of each other. We will prove this through a simple ablation experiment.

First, we compute the zero-shot judgments and ICR few-shot judgments for all test cases. Next, we identify any cases where the judgment has changed. Finally, we record semantic distances between each test case and the prompt demonstrations. We want to check if semantic distances have a certain relationship with judgment changes. 

\begin{figure}[h]
  \centering
  \subfloat[COLA]{\includegraphics[height = 0.4\linewidth]{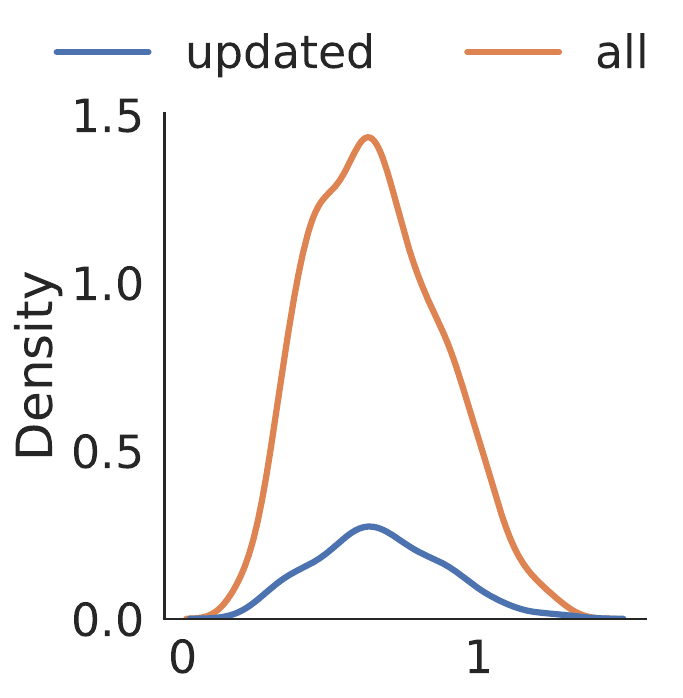}}
  \subfloat[MRPC]{\includegraphics[height = 0.4\linewidth]{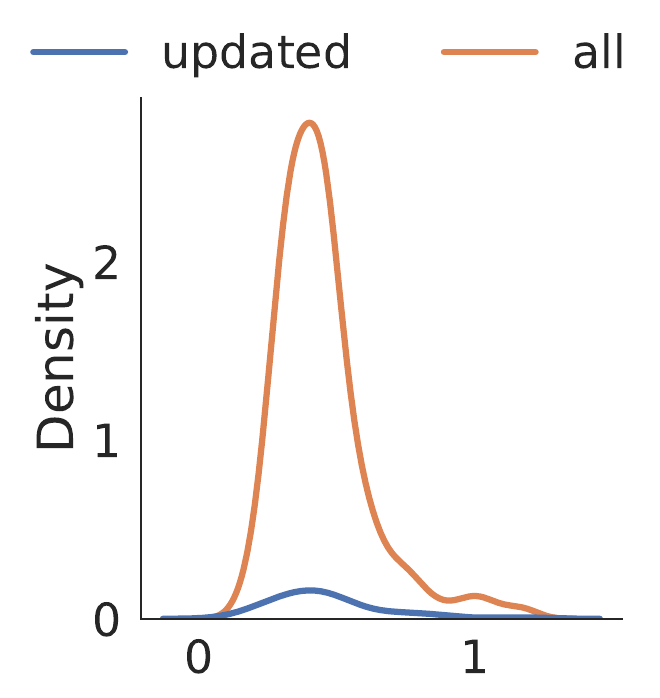}}
  \quad
  \subfloat[RTE]{\includegraphics[height = 0.4\linewidth]{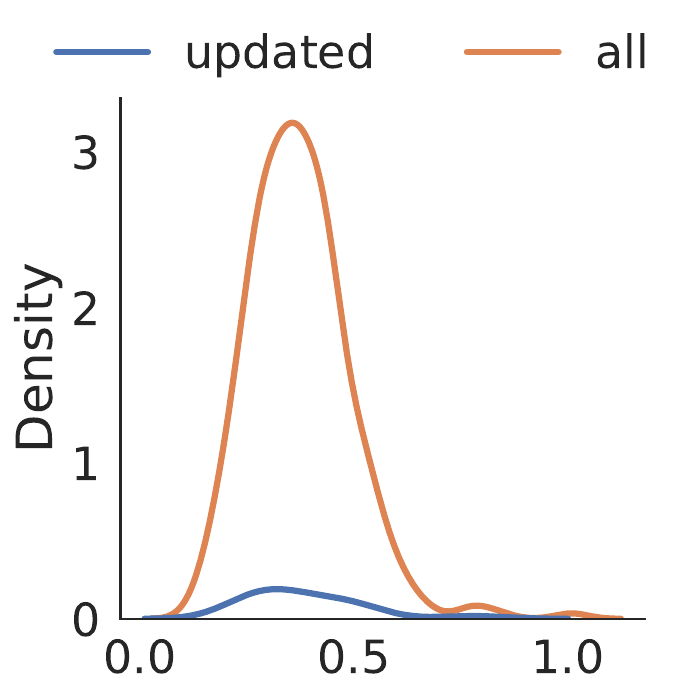}}
  \subfloat[WNLI]{\includegraphics[height = 0.4\linewidth]{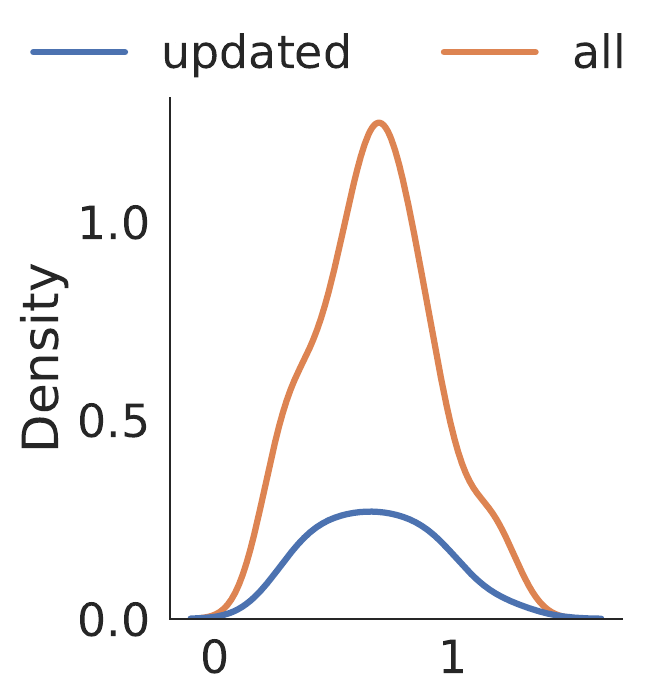}}
  \caption{Distribution of semantic distance between demonstrations and 1) all test cases 2) test cases whose judgment updated between zero-shot and few-shot prompts. There is no significant relationship between judgment updates and semantic distances.}
  \label{fig:ab3}
\end{figure}

Figure \ref{fig:ab3} illustrates the result. The distances of the modified cases are distributed evenly from the original distribution. This indicates that even if a test case is closer to one of the demonstrations, it does not have a higher chance of being correctly judged. This shows that the in-context influences of ICR prompts are independent of the semantic distances. As a result, the strategies mentioned above primarily rely on semantic features and do not leverage the distribution of the LLM.

\subsubsection{Case Study on ICR Improvement}

\begin{figure}[h]
  \centering
  \subfloat[$\psi$ distribution]{\includegraphics[height=0.32\linewidth]{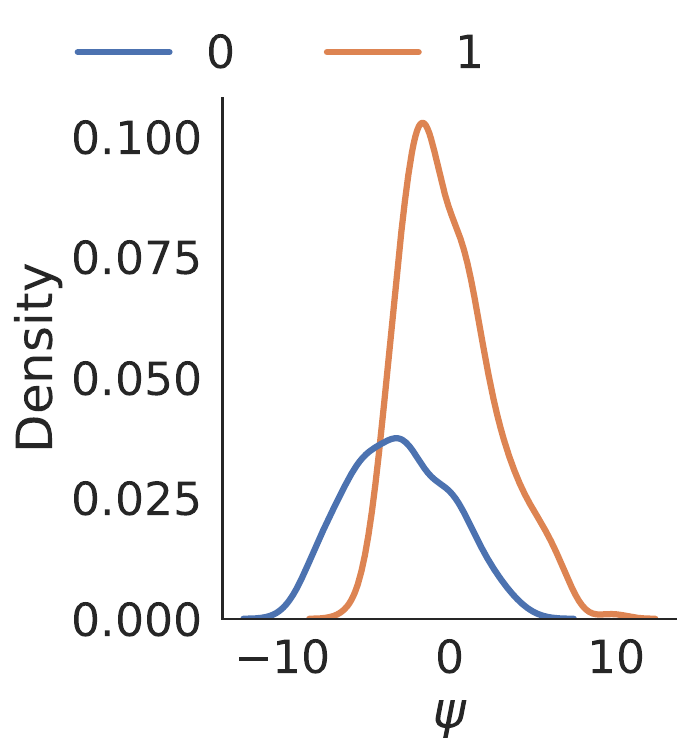}}
  \hfill
  \subfloat[Predictions before ICR]{\includegraphics[height=0.32\linewidth]{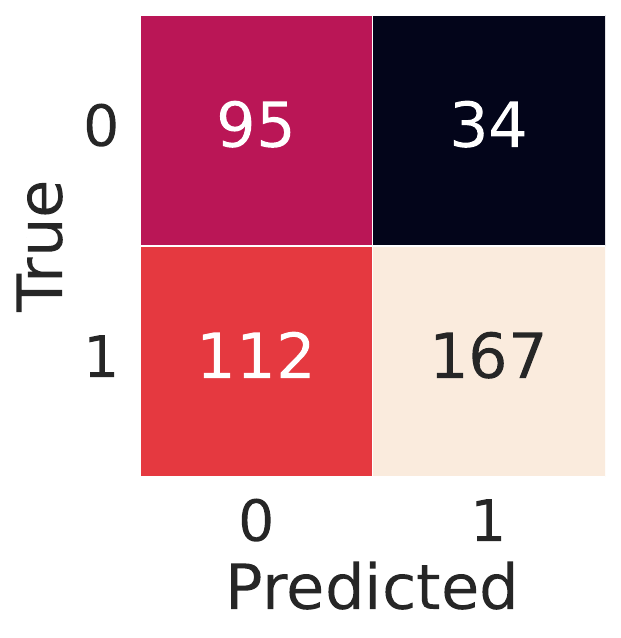}}
  \hfill
  \subfloat[Predictions after ICR]{\includegraphics[height=0.32\linewidth]{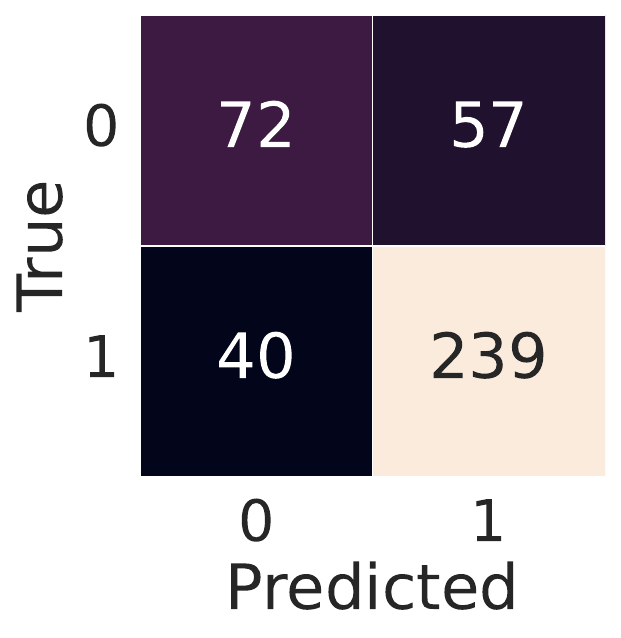}}
  \caption{Distribution of $\psi$ score on GLUE-MRPC's candidate pool, and the LLM predictions on test set before (after) apply ICR update. ICR sees more cases with label 1 being confidently misjudged. After refining the prompt with such cases, LLM's accuracy in judging case 1 is improved significantly.}
  \label{fig:ab4}
\end{figure}

To show how ICR bridges the discrepancy and further improves the task performance, we apply ICR on GLUE-MRPC, visualize the distribution of $\psi$ on demonstration candidates, and show the changes in LLM's prediction on test set. As results in Figure \ref{fig:ab4}, $\psi$ scores in the candidate pool show that cases with label 1 tend to be misjudged confidently. Therefore, ICR replaces part of the initial prompt with misjudged 1-labeled cases, bridging the gap between LLM's prediction and task mappings. This improves LLM's judging accuracy (especially on 1-labeled ones) significantly. 

\section{Conclusion}

We studied the problem of selecting demonstrations for in-context learning.
To tackle the limitations of existing methods, we proposed to leverage the discrepancies between LLM's knowledge and task expectations directly.
We proposed In-Context Reflection (ICR), a novel strategy that quantifies such discrepancy through misconfidence measurement.
Experiments showed ICR's prompt received an average 4\% gain on tasks.
Also, ICR received comparable performance when evaluated on tasks from the same task family, proving that ICR is robust.
Still, ICR has some limitations. More ICR iterations do not always improve the prompt. Also, performing ICR requires a fully labeled subset as a candidate pool. 
Future work can either investigate how to gain stable improvements with more iterations or try to build strategies jointly using discrepancy-based metrics with other measurements.

\bibliography{custom}
\bibliographystyle{acl_natbib}

\newpage
\appendix

\section{Task selection} \label{sec:unselected}

We only adopt part of the tasks under GLUE, TweetEval, and Ethos. One reason is that the GPT-3.5 backbone already performs high scores on the other tasks under uniform-sampling prompts. Therefore, it is hard to show our method's superiority under such tasks. Table \ref{tab:unselected} shows the result on GLUE-SST2 (SST2), Ethos-Sexual Orientation (E-SO), Ethos-Violence (E-V), Ethos-National Origin (E-NO), and Ethos-Disability (E-D). We didn't adopt GLUE-MNLI, GLUE-QNLI, GLUE-QQP, TweetEval-Emoji, and TweetEval-Sentiment as they contain too many test cases (9.8K, 5.8K, 293K, 50K, and 12.3K respectively) for us to afford. And we didn't adopt GLUE-STSB, as it is a numerical inference dataset, unsuitable for our general scope around classification tasks.

\begin{table}[h]
  \centering
  \caption{Performance of Uniform Sampling prompts on the unselected tasks. The performances are either too high or too low, and therefore we didn't select these tasks.}\label{tab:unselected}
  \begin{tabular}{c|ccccc}
    \toprule
    Task & SST2 & E-SO & E-V & E-NO & E-D \\ \midrule
    F1 & $95.1$ & $100$ & $91.0$ & $90.2$ & $89.7$ \\
    Acc & $95.0$ & $100$ & $94.3$ & $95.4$ & $90.8$ \\ \bottomrule
  \end{tabular}
\end{table}

\section{Datasets and Prompting Template}
\label{sec:app_dataset}

Sizes of all tasks in our experiments are shown in Table \ref{tab:dataset_size}. We follow the same data division as of Huggingface Datasets \citep{lhoest-etal-2021-datasets}, except for HateSpeech18, where we perform a custom division as only one data set is provided. The prompting templates used for each subtask are shown in Table \ref{tab:dataset_prompt}. We adopt a wide range of prompts from \citet{bach2022promptsource}.

\begin{table}[ht]
\centering
  \begin{centering}
    \caption{Dataset details}\label{tab:dataset_size}
    \begin{tabular}{p{0.5\linewidth}|cc}
    \hline
    Task & Train Size & Test Size \\ \hline
    GLUE-COLA & $8551$ & $1043$ \\ \hline
    GLUE-MRPC & $3668$ & $408$ \\ \hline
    GLUE-WNLI & $635$ & $71$ \\ \hline
    GLUE-RTE & $2490$ & $277$ \\ \hline
    Ethos-Religion & $346$ & $87$ \\ \hline
    Ethos-Race & $346$ & $87$ \\ \hline
    Ethos-Gender & $346$ & $87$ \\ \hline
    Ethos-Directed vs Generalized & $346$ & $87$ \\ \hline
    TweetEval-Hate & $9000$ & $2970$ \\ \hline
    TweetEval-Emotion & $3257$ & $374$ \\ \hline
    TweetEval-Irony & $2862$ & $955$ \\ \hline
    Hate Speech18 & $8755$ & $500$ \\ \hline
    Poem Sentiment & $892$ & $104$ \\ \hline
    \end{tabular}
  \end{centering}
\end{table}

\begin{table}[hb]
  \begin{centering}
    \caption{Prompt Examples}
    \label{tab:dataset_prompt}
    \begin{tabular}{p{0.2\linewidth}|p{0.8\linewidth}}
    \hline
    Task & Prompt Example \\ \hline
    GLUE-COLA & [Sentence] \textbackslash n Is this example grammatically correct and sensible?\textbackslash n [yes/no] \\ \hline
    GLUE-MRPC & Do the following two sentences mean the same thing?\textbackslash n [Sentence 1]\textbackslash n [Sentence 2]\textbackslash n [yes/no] \\ \hline
    GLUE-WNLI & Entailment means that the second sentence follows from the first sentence. Are the following two sentences an example of entailment?\textbackslash n [Sentence 1]\textbackslash n [Sentence 2]\textbackslash n [yes/no] \\ \hline
    GLUE-RTE & Does "[Sentence 1]" imply that "[Sentence 2]"? Please answer either yes or no.\textbackslash n [yes/no] \\ \hline
    Ethos & Text: [Sentence] \textbackslash n [Religious/Racial/Gender/Generalized] Hate: [yes/no] \\ \hline
    TweetEval-Hate & Text: [Sentence] \textbackslash n Hate: [yes/no] \\ \hline
    TweetEval-Emotion & [Sententce] \textbackslash n \textbackslash n What is the emotion of the text?\textbackslash n \textbackslash n Hint: anger, joy, optimism, sadness \textbackslash n [anger/joy/optimism/sadness] \\ \hline
    TweetEval-Irony & Is this tweet is ironic? \textbackslash n \textbackslash n [Sentence] \textbackslash n [yes/no] \\ \hline
    Hate Speech18 & Text: [Sentence] \textbackslash n Hate: [yes/no] \\ \hline
    Poem Sentiment & [Sentence] Is the sentiment the poet expresses for the poem negative, positive, neutral, or mixed? \textbackslash n \textbackslash n [negative/positive/neutral/mixed] \\ \hline
    \end{tabular}
  \end{centering}
\end{table}

\section{Hardware Details}

We use the Azure GPT-3.5-turbo-Instruct server to accomplish most experiments except tasks related to hate speech detection, where we use the original OpenAI API due to Azure's content filter policy. We use one Nvidia RTX A5000 GPU to hold pre-trained SBERT (as in KATE and AMBIG) or GPT2 (as in Topic).

\end{document}